\documentclass{article}
\usepackage{ijcai17}

\usepackage{times}

\usepackage{graphicx}
\usepackage{amsmath}
\usepackage{algorithm}
\usepackage{algpseudocode}
\usepackage{booktabs}
\usepackage{multirow}
\usepackage{amssymb}

\title{Generating Adversarial Malware Examples for Black-Box Attacks Based on GAN}
\author{Weiwei Hu \and Ying Tan\thanks{Prof. Ying Tan is the corresponding author.}\\
Key Laboratory of Machine Perception (MOE), and Department of Machine Intelligence\\
School of Electronics Engineering and Computer Science, Peking University, Beijing, 100871 China\\
\{weiwei.hu, ytan\}@pku.edu.cn}

\begin{document}

\maketitle

\begin{abstract}
Machine learning has been used to detect new malware in recent years, while malware authors have strong motivation to attack such algorithms.
Malware authors usually have no access to the detailed structures and parameters of the machine learning models used by malware detection systems, and therefore they can only perform black-box attacks.
This paper proposes a generative adversarial network (GAN) based algorithm named MalGAN to generate adversarial malware examples, which are able to bypass black-box machine learning based detection models.
MalGAN uses a substitute detector to fit the black-box malware detection system. A generative network is trained to minimize the generated adversarial examples' malicious probabilities predicted by the substitute detector.
The superiority of MalGAN over traditional gradient based adversarial example generation algorithms is that MalGAN is able to decrease the detection rate to nearly zero and make the retraining based defensive method against adversarial examples hard to work.
\end{abstract}

\section{Introduction}
In recent years, many machine learning based algorithms have been proposed to detect malware, which extract features from programs and use a classifier to classify programs between benign programs and malware. For example, Schultz et al. proposed to use DLLs, APIs and strings as features for classification \cite{schultz2001data}, while Kolter et al. used byte level N-Gram as features \cite{kolter2004learning,kolter2006learning}.

Most researchers focused their efforts on improving the detection performance (e.g. true positive rate, accuracy and AUC) of such algorithms, but ignored the robustness of these algorithms. Generally speaking, the propagation of malware will benefit malware authors. Therefore, malware authors have sufficient motivation to attack malware detection algorithms.

Many machine learning algorithms are very vulnerable to intentional attacks. Machine learning based malware detection algorithms cannot be used in real-world applications if they are easily to be bypassed by some adversarial techniques.

Recently, adversarial examples of deep learning models have attracted the attention of many researchers. Szegedy et al. added imperceptible perturbations to images to maximize a trained neural network's classification errors, making the network unable to classify the images correctly \cite{szegedy2013intriguing}. The examples after adding perturbations are called adversarial examples. Goodfellow et al. proposed a gradient based algorithm to generate adversarial examples \cite{goodfellow2014explaining}. Papernot et al. used the Jacobian matrix to determine which features to modify when generating adversarial examples \cite{papernot2016limitations}. The Jacobian matrix based approach is also a kind of gradient based algorithm.

Grosse et al. proposed to use the gradient based approach to generate adversarial Android malware examples \cite{grosse2016adversarial}. The adversarial examples are used to fool a neural network based malware detection model. They assumed that attackers have full access to the parameters of the malware detection model. For different sizes of neural networks, the misclassification rates after adversarial crafting range from 40\% to 84\%. 

In some cases, attackers have no access to the architecture and weights of the neural network to be attacked; the target model is a black box to attackers. Papernot et al. used a substitute neural network to fit the black-box neural network and then generated adversarial examples according to the substitute neural network \cite{papernot2016practical}. They also used a substitute neural network to attack other machine learning algorithms such as logistic regression, support vector machines, decision trees and nearest neighbors \cite{papernot2016transferability}. Liu et al. performed black-box attacks without a substitute model \cite{liu2016delving}, based on the principle that adversarial examples can transfer among different models \cite{szegedy2013intriguing}.

Machine learning based malware detection algorithms are usually integrated into antivirus software or hosted on the cloud side, and therefore they are black-box systems to malware authors. It is hard for malware authors to know which classifier a malware detection system uses and the parameters of the classifier.

However, it is possible to figure out what features a malware detection algorithm uses by feeding some carefully designed test cases to the black-box algorithm. For example, if a malware detection algorithm uses static DLL or API features from the import directory table or the import lookup tables of PE programs \cite{microsoft2016pe}, malware authors can manually modify some DLL or API names in the import directory table or the import lookup tables. They can modify a benign program's DLL or API names to malware's DLL or API names, and vice versa. If the detection results change after most of the modifications, they can judge that the malware detection algorithm uses DLL or API features. Therefore, in this paper we assume that malware authors are able to know what features a malware detection algorithm uses, but know nothing about the machine learning model.

Existing algorithms mainly use gradient information and hand-crafted rules to transform original samples into adversarial examples. This paper proposes a generative neural network based approach which takes original samples as inputs and outputs adversarial examples. The intrinsic non-linear structure of neural networks enables them to generate more complex and flexible adversarial examples to fool the target model.

The learning algorithm of our proposed model is inspired by generative adversarial networks (GAN) \cite{goodfellow2014generative}. In GAN, a discriminative model is used to distinguish between generated samples and real samples, and a generative model is trained to make the discriminative model misclassify generated samples as real samples. GAN has shown good performance in generating realistic images\cite{mirza2014conditional,denton2015deep}.

The proposed model in this paper is named as MalGAN, which generates adversarial examples to attack black-box malware detection algorithms. A substitute detector is trained to fit the black-box malware detection algorithm, and a generative network is used to transform malware samples into adversarial examples. Experimental results show that almost all of the adversarial examples generated by MalGAN successfully bypass the detection algorithms and MalGAN is very flexible to fool further defensive methods of detection algorithms.

\section{Architecture of MalGAN}
\subsection{Overview}
\begin{figure*}[htp]
    \begin{center}
    \graphicspath{{img/}}
    \includegraphics[width = 5.5in]{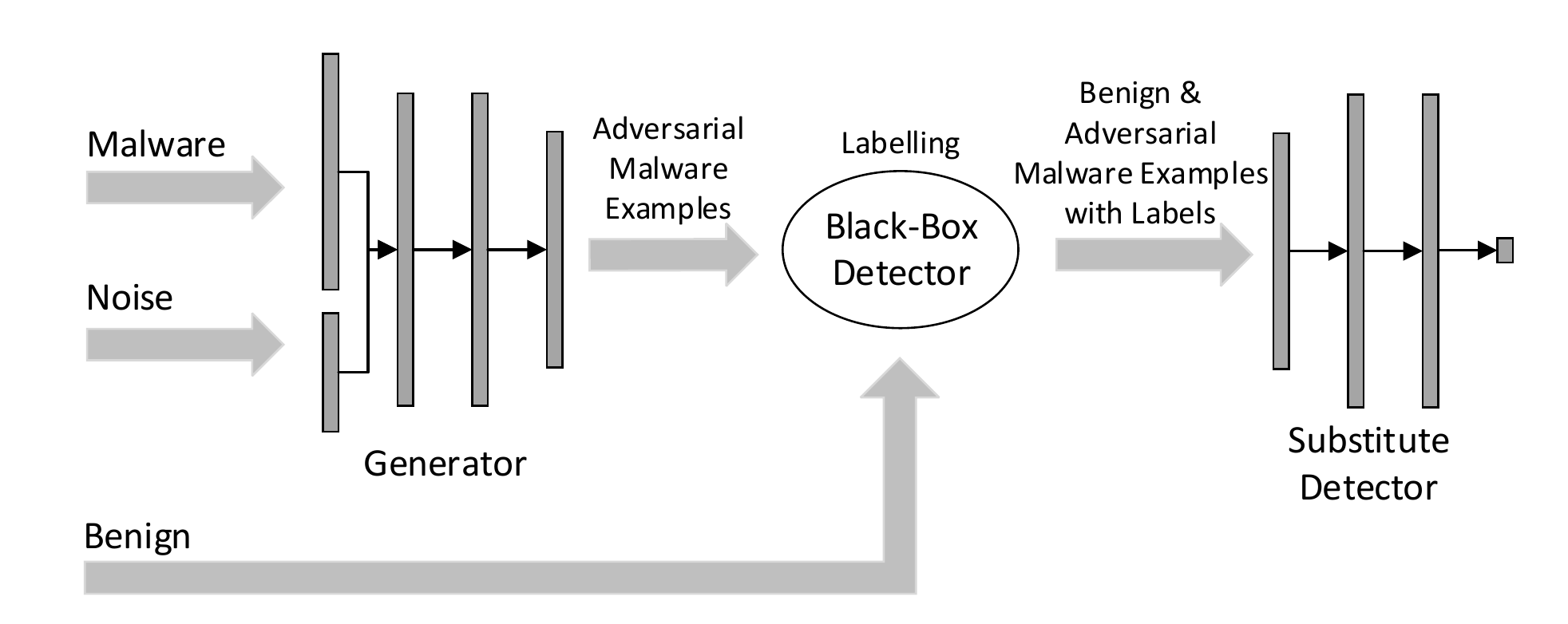}
    \caption{The architecture of MalGAN.}
    \label{fig:malgan}
    \end{center}
\end{figure*}

The architecture of proposed MalGAN is shown in Figure \ref{fig:malgan}.

The black-box detector is an external system which adopts machine learning based malware detection algorithms. We assume that the only thing malware authors know about the black-box detector is what kind of features it uses. Malware authors do not know what machine learning algorithm it uses and do not have access to the parameters of the trained model. Malware authors are able to get the detection results of their programs from the black-box detector. The whole model contains a generator and a substitute detector, which are both feed-forward neural networks. The generator and the substitute detector work together to attack a machine learning based black-box malware detector.

In this paper we only generate adversarial examples for binary features, because binary features are widely used by malware detection researchers and are able to result in high detection accuracy. Here we take API feature as an example to show how to represent a program. If $M$ APIs are used as features, an $M$-dimensional feature vector is constructed for a program. If the program calls the $d$-th API, the $d$-th feature value is set to 1, otherwise it is set to 0.

The main difference between this model and existing algorithms is that the adversarial examples are dynamically generated according to the feedback of the black-box detector, while most existing algorithms use static gradient based approaches to generate adversarial examples.

The probability distribution of adversarial examples from MalGAN is determined by the weights of the generator. To make a machine learning algorithm effective, the samples in the training set and the test set should follow the same probability distribution or similar probability distributions. While the generator can change the probability distribution of adversarial examples to make it far from the probability distribution of the black-box detector's training set. In this case the generator has sufficient opportunity to lead the black-box detector to misclassify malware as benign.

\subsection{Generator}
The generator is used to transform a malware feature vector into its adversarial version. It takes the concatenation of a malware feature vector $\boldsymbol{m}$ and a noise vector $\boldsymbol{z}$ as input. $\boldsymbol{m}$ is a $M$-dimensional binary vector. Each element of $\boldsymbol{m}$ corresponds to the presence or absence of a feature. $\boldsymbol{z}$ is a $Z$-dimensional vector, where $Z$ is a hyper-parameter. Each element of $\boldsymbol{z}$ is a random number sampled from a uniform distribution in the range $[0, 1)$. The effect of $\boldsymbol{z}$ is to allow the generator to generate diverse adversarial examples from a single malware feature vector.

The input vector is fed into a multi-layer feed-forward neural network with weights $\theta_g$. The output layer of this network has $M$ neurons and the activation function used by the last layer is sigmoid which restricts the output to the range $(0, 1)$. The output of this network is denoted as $\boldsymbol{o}$. Since malware feature values are binary, binarization transformation is applied to $\boldsymbol{o}$ according to whether an element is greater than 0.5 or not, and this process produces a binary vector $\boldsymbol{o'}$.

When generating adversarial examples for binary malware features we only consider to add some irrelevant features to malware. Removing a feature from the original malware may crack it. For example, if the ``WriteFile" API is removed from a program, the program is unable to perform normal writing function and the malware may crack. The non-zero elements of the binary vector $\boldsymbol{o'}$ act as the irrelevant features to be added to the original malware. The final generated adversarial example can be expressed as $\boldsymbol{m'}=\boldsymbol{m}|\boldsymbol{o'}$ where ``$|$" is element-wise binary OR operation.

$\boldsymbol{m'}$ is a binary vector, and therefore the gradients are unable to back propagate from the substitute detector to the generator. A smooth function $G$ is defined to receive gradient information from the substitute detector, as shown in Formula \ref{equ:g}.

\begin{equation}
\label{equ:g}
G_{\theta_g}(\boldsymbol{m},\boldsymbol{z}) = \max \left(\boldsymbol{m}, \boldsymbol{o} \right) .
\end{equation}

$\max \left( \cdotp , \cdotp \right)$ represents element-wise max operation. If an element of $\boldsymbol{m}$ has the value 1, the corresponding result of $G$ is also 1, which is unable to back propagate the gradients. If an element of $\boldsymbol{m}$ has the value 0, the result of $G$ is the neural network's real number output in the corresponding dimension, and gradient information is able to go through. It can be seen that $\boldsymbol{m'}$ is actually the binarization transformed version of  $G_{\theta_g}(\boldsymbol{m},\boldsymbol{z})$.

\subsection{Substitute Detector}
Since malware authors know nothing about the detailed structure of the black-box detector, the substitute detector is used to fit the black-box detector and provides gradient information to train the generator.

The substitute detector is a multi-layer feed-forward neural network with weights $\theta_d$ which takes a program feature vector $\boldsymbol{x}$ as input. It classifies the program between benign program and malware. We denote the predicted probability that $\boldsymbol{x}$ is malware as $D_{\theta_d}(\boldsymbol{x})$.

The training data of the substitute detector consist of adversarial malware examples from the generator, and benign programs from an additional benign dataset collected by malware authors. The ground-truth labels of the training data are not used to train the substitute detector. The goal of the substitute detector is to fit the black-box detector. The black-box detector will detect this training data first and output whether a program is benign or malware. The predicted labels from the black-box detector are used by the substitute detector.

\section{Training MalGAN}
To train MalGAN malware authors should collect a malware dataset and a benign dataset first.

The loss function of the substitute detector is defined in Formula \ref{equ:d-loss}.

\begin{equation}
\label{equ:d-loss}
\begin{split}
{L_D} = & - {\mathbb{E}_{\boldsymbol{x}\in{BB_{Benign}}}}\log\left( {1 - D_{\theta_d}(\boldsymbol{x})} \right)\\
 &- {\mathbb{E}_{\boldsymbol{x}\in{BB_{Malware}}}}\log {D_{\theta_d}\left( \boldsymbol{x} \right)}.
\end{split}
\end{equation}

$BB_{Benign}$ is the set of programs that are recognized as benign by the black-box detector, and $BB_{Malware}$ is the set of programs that are detected as malware by the black-box detector.

To train the substitute detector, $L_D$ should be minimized with respect to the weights of the substitute detector.

The loss function of the generator is defined in Formula \ref{equ:g-loss}.

\begin{equation}
\label{equ:g-loss}
{L_G} = {\mathbb{E}_{\boldsymbol{m}\in{S_{Malware}},\boldsymbol{z}\sim{\boldsymbol{p}_{{\rm{uniform}}[0,1)}}}}\log {D_{\theta_d}\left( {G_{\theta_g}\left( {\boldsymbol{m},\boldsymbol{z}} \right)} \right)}.
\end{equation}

$S_{Malware}$ is the actual malware dataset, not the malware set labelled by the black-box detector. $L_G$ is minimized with respect to the weights of the generator.

Minimizing $L_G$ will reduce the predicted malicious probability of malware and push the substitute detector to recognize malware as benign. Since the substitute detector tries to fit the black-box detector, the training of the generator will further fool the black-box detector.

The whole process of training MalGAN is shown in Algorithm \ref{alg:training}.

\begin{algorithm}
\caption{The Training Process of MalGAN}
\label{alg:training}
\begin{algorithmic}[1]
\While{not converging}
    \State \label{step:m}Sample a minibatch of malware $\boldsymbol{M}$
    \State Generate adversarial examples $\boldsymbol{M'}$ from the generator for $\boldsymbol{M}$
    \State \label{step:b}Sample a minibatch of benign programs $\boldsymbol{B}$
    \State Label $\boldsymbol{M'}$ and $\boldsymbol{B}$ using the black-box detector
    \State Update the substitute detector's weights $\theta_d$ by descending along the gradient ${\nabla _{{\theta _d}}}{L_D}$
    \State Update the generator's weights $\theta_g$ by descending along the gradient ${\nabla _{{\theta _g}}}{L_G}$
\EndWhile
\end{algorithmic}
\end{algorithm}

In line \ref{step:m} and line \ref{step:b}, different sizes of minibatches are used for malware and benign programs. The ratio of $\boldsymbol{M}$'s size to $\boldsymbol{B}$'s size is the same as the ratio of the malware dataset's size to the benign dataset's size.

\section{Experiments}
\subsection{Experimental Setup}
The dataset used in this paper was crawled from a program sharing website\footnote{https://malwr.com/}. We downloaded 180 thousand programs from this website and about 30\% of them are malware. API features are used in this paper. An 160-dimensional binary feature vector is construct for each program, based on 160 system level APIs.

In order to validate the transferability of adversarial examples generated by MalGAN, we tried several different machine learning algorithms for the black-box detector. The used classifiers include random forest (RF), logistic regression (LR), decision trees (DT), support vector machines (SVM), multi-layer perceptron (MLP), and a voting based ensemble of these classifiers (VOTE).

We adopted two ways to split the dataset. The first splitting way regards 80\% of the dataset as the training set and the remaining 20\% as the test set. MalGAN and the black-box detector share the same training set. MalGAN further picks out 25\% of the training data as the validation set and uses the remaining training data to train the neural networks. Some black-box classifiers such as MLP also need a validation set for early stopping. The validation set of MalGAN cannot be used for the black-box detector since malware authors and antivirus vendors do not communicate on how to split dataset. Splitting validation set for the black-box detector should be independent of MalGAN; MalGAN and the black-box detector should use different random seeds to pick out the validation data.

The second splitting way picks out 40\% of the dataset as the training set for MalGAN, picks out another 40\% of the dataset as the training set for the black-box detector, and uses the remaining 20\% of the dataset as the test set.

In real-world scenes the training data collected by the malware authors and the antivirus vendors cannot be the same. However, their training data will overlap with each other if they collect data from public sources. In this case the actual performance of MalGAN will be between the performances of the two splitting ways.

Adam \cite{kingma2014adam} was chosen as the optimizer. We tuned the hyper-parameters on the validation set. 10 was chosen as the dimension of the noise vector $\boldsymbol{z}$. The generator's layer size was set to 170-256-160, the substitute detector's layer size was set to 160-256-1, and the learning rate 0.001 was used for both the generator and the substitute detector. The maximum number of epochs to train MalGAN was set to 100. The epoch with the lowest detection rate on the validation set is finally chosen to test the performance of MalGAN.

\subsection{Experimental Results}
We first analyze the case where MalGAN and the black-box detector use the same training set. For malware detection, the true positive rate (TPR) means the detection rate of malware. After adversarial attacks, the reduction in TPR can reflect how many malware samples successfully bypass the detection algorithm. TPR on the training set and the test set of original samples and adversarial examples is shown in Table \ref{tab:samedata}.

\begin{table}[htbp]
  \centering
  \caption{True positive rate (in percentage) on original samples and adversarial examples when MalGAN and the black-box detector are trained on the same training set. ``Adver." represents adversarial examples.}
    \begin{tabular}{llllll}
    \toprule
    & \multicolumn{2}{l}{Training Set} & & \multicolumn{2}{l}{Test Set}  \\
    \cline{2-3}\cline{5-6}
          & Original& Adver.& & Original & Adver. \\
          \midrule
    RF    & 97.62 & 0.20 & & 95.38 & 0.19 \\
    LR    & 92.20 & 0.00 & & 92.27 & 0.00 \\
    DT    & 97.89 & 0.16 & & 93.98 & 0.16 \\
    SVM   & 93.11 & 0.00 & & 93.13 & 0.00 \\
    MLP   & 95.11 & 0.00     & & 94.89 & 0.00 \\
    VOTE & 97.23 & 0.00     & & 95.64 & 0.00 \\
    \bottomrule
    \end{tabular}%
  \label{tab:samedata}%
\end{table}%

For random forest and decision trees, the TPRs on adversarial examples range from 0.16\% to 0.20\% for both the training set and the test set, while the TPRs on the original samples are all greater than 93\%. When using other classifiers as the black-box detector, MalGAN is able to decrease the TPR on generated adversarial examples to zero for both the training set and the test set. That is to say, for all of the backend classifiers, the black-box detector can hardly detect any malware generated by the generator. The proposed model has successfully learned to bypass these machine learning based malware detection algorithms.

The structures of logistic regression and support vector machines are very similar to neural networks and MLP is actually a neural network. Therefore, the substitute detector is able to fit them with a very high accuracy. This is why MalGAN can achieve zero TPR for these classifiers. While random forest and decision trees have quite different structures from neural networks so that MalGAN results in non-zero TPRs. The TPRs of random forest and decision trees on adversarial examples are still quite small, which means the neural network has enough capacity to represent other models with quite different structures. The voting of these algorithms also achieves zero TPR. We can conclude that the classifiers with similar structures to neural networks are in the majority during voting.

\begin{figure}[htp]
    \begin{center}
    \graphicspath{{img/}}
    \includegraphics[width = 3.0in]{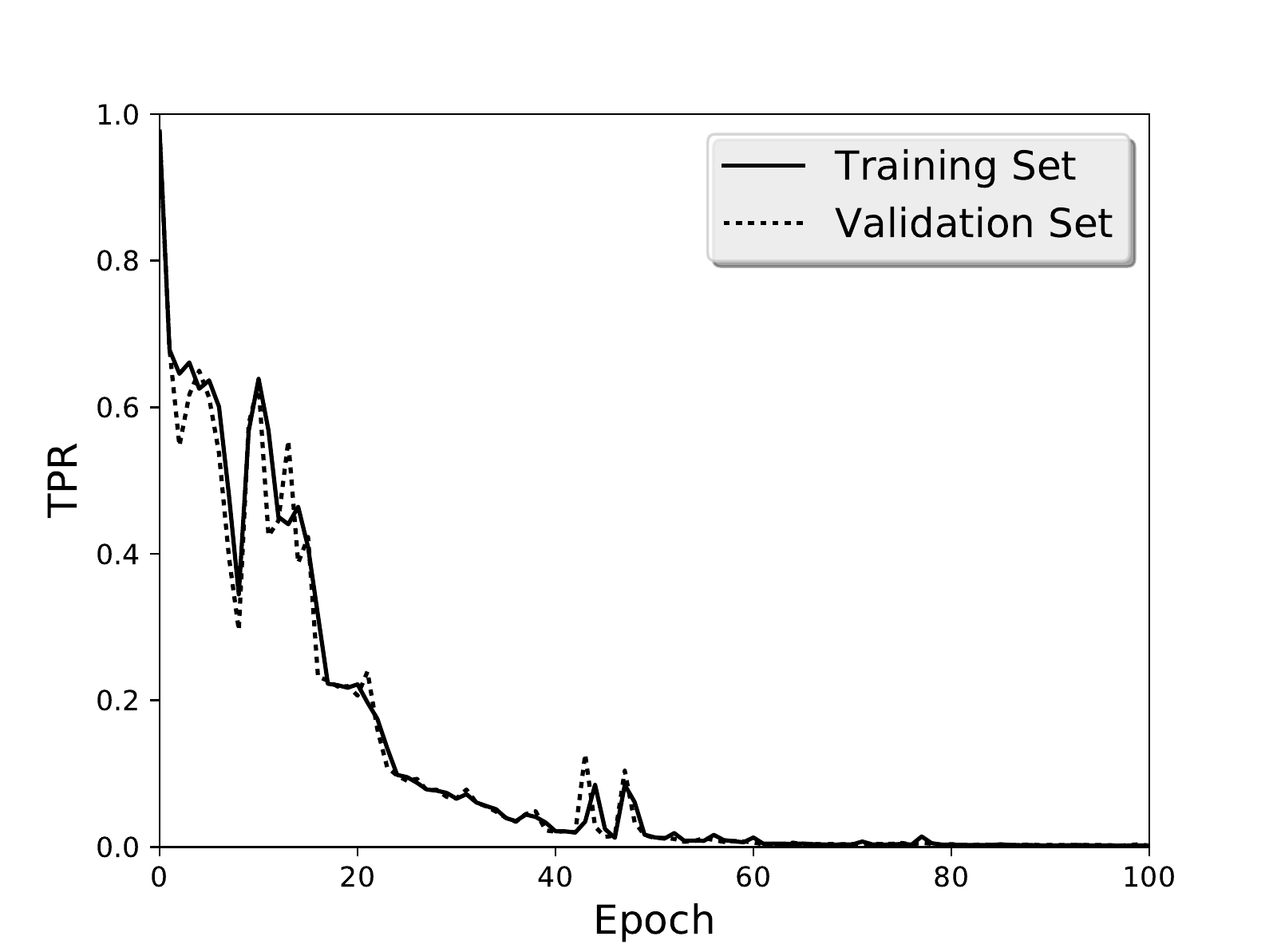}
    \caption{The change of the true positive rate on the training set and the validation set over time. Random forest is used as the black-box detector here. The vertical axis represents the true positive rate while the horizontal axis represents epoch.}
    \label{fig:rftpr}
    \end{center}
\end{figure}

The convergence curve of TPR on the training set and the validation set during the training process of MalGAN is shown in Figure \ref{fig:rftpr}. The black-box detector used here is random forest, since random forest performs very well in Table \ref{tab:samedata}.

TPR converges to about zero near the 40th epoch, but the convergence curve is a bit shaking, not a smooth one. This curve reflects the fact that the training of GAN is usually unstable. How to stabilize the training of GAN have attracted the attention of many researchers \cite{radford2015unsupervised,salimans2016improved,arjovsky2017towards}.

Now we will analyze the results when MalGAN and the black-box detector are trained on different training sets. Fitting the black-box detector trained on a different dataset is more difficult for the substitute detector. The experimental results are shown in Table \ref{tab:differentdatatpr}.

\begin{table}[htbp]
  \centering
  \caption{True positive rate (in percentage) on original samples and adversarial examples when MalGAN and the black-box detector are trained on different training sets. ``Adver." represents adversarial examples.}
    \begin{tabular}{llllll}
    \toprule
    & \multicolumn{2}{l}{Training Set} & & \multicolumn{2}{l}{Test Set}  \\
    \cline{2-3}\cline{5-6}
          & Original& Adver.& & Original & Adver. \\
          \midrule
    RF    & 95.10 & 0.71 & & 94.95 & 0.80 \\
    LR    & 91.58 & 0.00 & & 91.81 & 0.01 \\
    DT    & 91.92 & 2.18 & & 91.97 & 2.11 \\
    SVM   & 92.50 & 0.00     & & 92.78 & 0.00 \\
    MLP   & 94.32 & 0.00 & & 94.40 & 0.00 \\
    VOTE & 94.30 & 0.00 & & 94.45 & 0.00 \\
    \bottomrule
    \end{tabular}%
  \label{tab:differentdatatpr}%
\end{table}%

For SVM, MLP and VOTE, TPR reaches zero, and TPR of LR is nearly zero. These results are very similar to Table \ref{tab:samedata}. TPRs of random forest and decision trees on adversarial examples become higher compared with the case where MalGAN and the black-box detector use the same training data. For decision trees the TPRs rise to 2.18\% and 2.11\% on the training set and the test set respectively. However, 2\% is still a very small number and the black-box detector will still miss to detect most of the adversarial malware examples. It can be concluded that MalGAN is still able to fool the black-box detector even trained on a different training set.

\subsection{Comparison with the Gradient based Algorithm to Generate Adversarial Examples}
Existing algorithms of generating adversarial examples are mainly for images. The difference between image and malware is that image features are continuous while malware features are binary.

Grosse et al. modified the traditional gradient based algorithm to generate binary adversarial malware examples \cite{grosse2016adversarial}. They did not regard the malware detection algorithm as a black-box system and assumed that malware authors have full access to the architecture and the weights of the neural network based malware detection model. The misclassification rates of adversarial examples range from 40\% to 84\% under different hyper-parameters. This gradient based approach under white-box assumption is unable to generate adversarial examples with zero TPR, while MalGAN produces nearly zero TPR with a harder black-box assumption.

Their algorithm uses an iterative approach to generate adversarial malware examples. At each iteration the algorithm finds the feature with the maximum likelihood to change the malware's label from malware to benign. The algorithm modifies one feature at each iteration, until the malware is successfully classified as a benign program or there are no features available to be modified.

We tried to migrate this algorithm to attack a random forest based black-box detection algorithm. A substitute neural network is trained to fit the black-box random forest. Adversarial malware examples are generated based on the gradient information of the substitute neural network.

\begin{figure}[htp]
    \begin{center}
    \graphicspath{{img/}}
    \includegraphics[width = 3.0in]{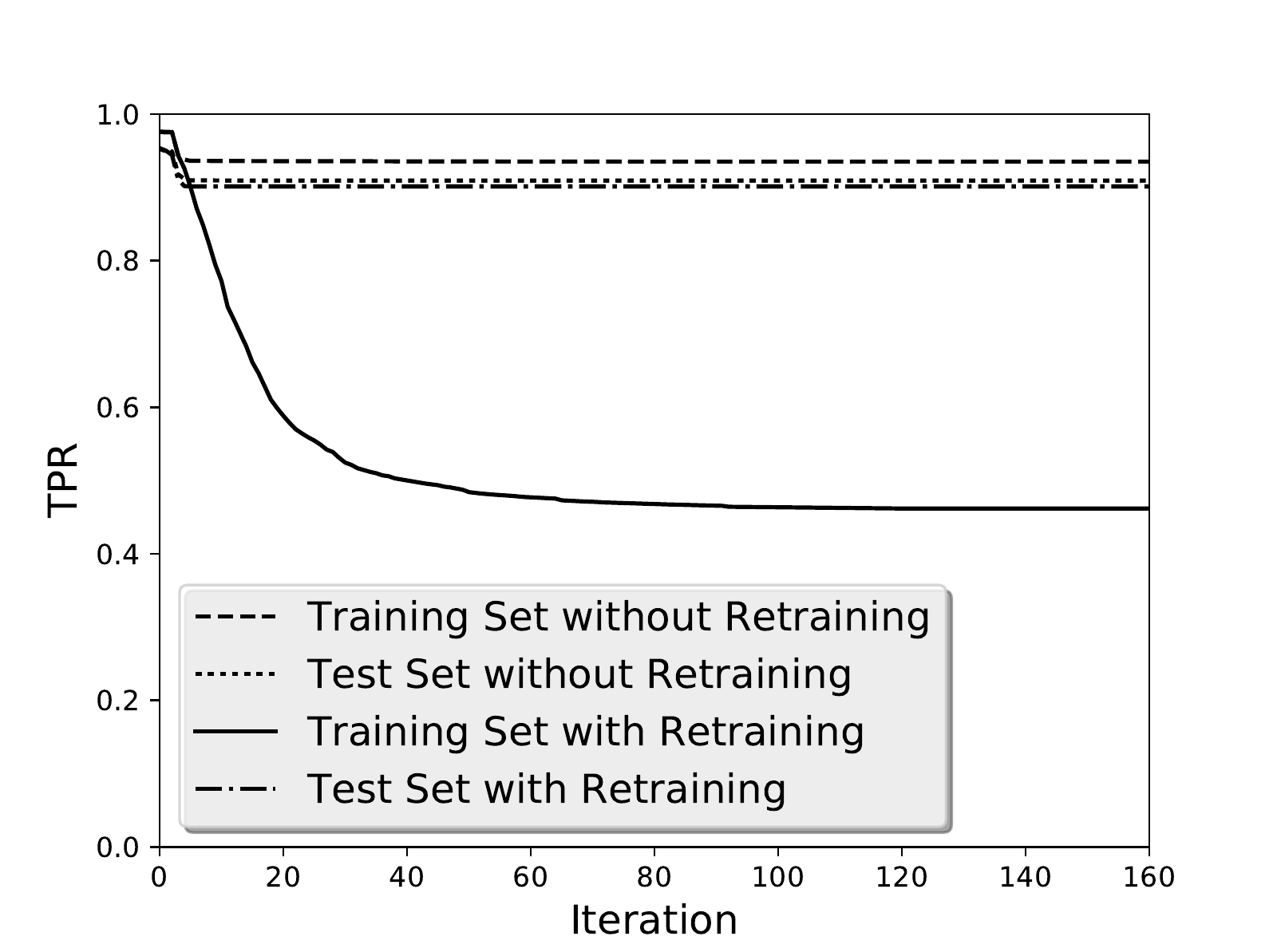}
    \caption{True positive rate on the adversarial examples over the iterative process when using the algorithm proposed by Grosse et al..}
    \label{fig:grad}
    \end{center}
\end{figure}

TPR on the adversarial examples over the iterative process is shown in Figure \ref{fig:grad}. Please note that at each iteration not all of the malware samples are modified. If a malware sample has already been classified as a benign program at previous iterations or there are no modifiable features, the algorithm will do nothing on the malware sample at this iteration.

On the training set and the test set, TPR converges to 93.52\% and 90.96\% respectively. In this case the black-box random forest is able to detect most of the adversarial examples. The substitute neural network is trained on the original training set, while after several iterations the probability distribution of adversarial examples will become quite different from the probability distribution of the original training set. Therefore, the substitute neural network cannot approximate the black-box random forest well on the adversarial examples. In this case the adversarial examples generated from the substitute neural network are unable to fool the black-box random forest.

In order to fit the black-box random forest more accurately on the adversarial examples, we tried to retraining the substitute neural network on the adversarial examples. At each iteration, the current generated adversarial examples from the whole training set are used to retrain the substitute neural network. As shown in Figure \ref{fig:grad}, the retraining approach make TPR converge to 46.18\% on the training set, which means the black-box random forest can still detect about half of the adversarial examples. However, the retrained model is unable to generalize to the test set, sine the TPR on the test set converges to 90.12\%. The odd probability distribution of these adversarial examples limits the generalization ability of the substitute neural network.

MalGAN uses a generative network to transform original samples into adversarial samples. The neural network has enough representation ability to perform complex transformations, making MalGAN able to result in nearly zero TPR on both the training set and the test set. While the representation ability of the gradient based approach is too limited to generate high-quality adversarial examples.

\subsection{Retraining the Black-Box Detector}
Several defensive algorithms have been proposed to deal with adversarial examples. Gu et al. proposed to use auto-encoders to map adversarial samples to clean input data \cite{gu2014towards}. An algorithm named defensive distillation was proposed by Papernot et al. to weaken the effectiveness of adversarial perturbations \cite{papernot2016distillation}. Li et al. found that adversarial retraining can boost the robustness of machine learning algorithms \cite{li2016general}. Chen et al. compared these defensive algorithms and concluded that retraining is a very effective way to defend against adversarial examples, and is robust even against repeated attacks \cite{chen2016evaluation}.

In this section we will analyze the performance of MalGAN under the retraining based defensive approach. If antivirus vendors collect enough adversarial malware examples, the can retrain the black-box detector on these adversarial examples in order to learn their patterns and detect them. Here we only use random forest as the black-box detector due to its good performance. After retraining the black-box detector it is able to detect all adversarial examples, as shown in the middle column of Table \ref{tab:retraintpr}.

\begin{table}[htbp]
  \centering
  \caption{True positive rate (in percentage) on the adversarial examples after the black-box detector is retrained.}
    \begin{tabular}{lp{2.5cm}p{2.5cm}}
    \toprule
       & Before Retraining MalGAN & After Reraining MalGAN\\
    \midrule
    Training set & 100 & 0 \\
    Test set & 100 & 0 \\
    \bottomrule
    \end{tabular}%
  \label{tab:retraintpr}%
\end{table}%

However, once antivirus vendors release the updated black-box detector publicly, malware authors will be able to get a copy of it and retrain MalGAN to attack the new black-box detector. After this process the black-box detector can hardly detect any malware again, as shown in the last column of Table \ref{tab:retraintpr}. We found that reducing TPR from 100\% to 0\% can be done within one epoch during retraining MalGAN. We alternated retraining the black-box detector and retraining MalGAN for ten times. The results are the same as Table \ref{tab:retraintpr} for the ten times.

To retrain the black-box detector antivirus vendors have to collect enough adversarial examples. It is a long process to collect a large number of malware samples and label them. Adversarial malware examples have enough time to propagate before the black-box detector is retrained and updated. Once the black-box detector is updated, malware authors will attack it immediately by retraining MalGAN and our experiments showed that retraining takes much less time than the first-time training. After retraining MalGAN, new adversarial examples remain undetected. This dynamic adversarial process lands antivirus vendors in a passive position. Machine learning based malware detection algorithms can hardly work in this case.


\section{Conclusions}
This paper proposed a novel algorithm named MalGAN to generate adversarial examples from a machine learning based black-box malware detector. A neural network based substitute detector is used to fit the black-box detector. A generator is trained to generate adversarial examples which are able to fool the substitute detector. Experimental results showed that the generated adversarial examples are able to effectively bypass the black-box detector.

Adversarial examples' probability distribution is controlled by the weights of the generator. Malware authors are able to frequently change the probability distribution by retraining MalGAN, making the black-box detector cannot keep up with it, and unable to learn stable patterns from it. Once the black-box detector is updated malware authors can immediately crack it. This process making machine learning based malware detection algorithms unable to work.

\section*{Acknowledgments}
This work was supported by the Natural Science Foundation of China (NSFC) under grant no. 61375119 and the Beijing Natural Science Foundation under grant no. 4162029, and partially supported by National Key Basic Research Development Plan (973 Plan) Project of China under grant no. 2015CB352302.

\bibliographystyle{named}
\bibliography{ijcai17}

\end{document}